\documentclass[letterpaper, 10 pt, conference]{ieeeconf}  
\usepackage{FG2025}
\usepackage{graphicx}
\usepackage{url}

\IEEEoverridecommandlockouts                              
\usepackage{graphics} 
\usepackage{epsfig} 
\usepackage{mathptmx} 
\usepackage{times} 
\usepackage{amsmath} 
\usepackage{amssymb}  

\FGfinalcopy 

\def\FGPaperID{****} 

\title{\LARGE \bf
Expanding on the BRIAR Dataset: A Comprehensive Whole Body Biometric Recognition Resource at Extreme Distances and Real-World Scenarios (Collections 1-4)
}

\author{\parbox{16cm}{\centering
    {\large Gavin Jager, David Cornett III, Gavin Glenn, Deniz Aykac, Christi Johnson, Robert Zhang, Ryan Shivers, David Bolme, Laura Davies, Scott Dolvin, Nell Barber, Joel Brogan, Nick Burchfield, Carl Dukes, Andrew Duncan, Regina Ferrell, Austin Garrett, Jim Goddard, Jairus Hines, Bart Murphy, Sean Pharris, Brandon Stockwell, Leanne Thompson and Matthew Yohe}\\
    {\normalsize
    Oak Ridge National Laboratory\\}}
    \thanks{IARPA}
}

\begin{document}

\ifFGfinal
\thispagestyle{empty}
\pagestyle{empty}
\else
\author{Anonymous FG2025 submission\\ Paper ID \FGPaperID \\}
\pagestyle{plain}
\fi
\maketitle

\begin{abstract}

The state-of-the-art in biometric recognition algorithms and operational systems has advanced quickly in recent years providing high accuracy and robustness in more challenging collection environments and consumer applications. However, the technology still suffers greatly when applied to non-conventional settings such as those seen when performing identification at extreme distances or from elevated cameras on buildings or mounted to UAVs. This paper summarizes an extension to the largest dataset currently focused on addressing these operational challenges, and describes its composition as well as methodologies of collection, curation, and annotation.

\end{abstract}

\section{INTRODUCTION}
\label{sec:intro}

The Biometric Recognition at Altitude and Range (BRIAR) Program is
a US Government sponsored
initiative to advance the state of the art of biometric recognition under challenging conditions. The overarching goal is to develop end-to-end software systems capable of overcoming severe atmospheric distortion and difficult imaging conditions, perform person detection and tracking, and fuse multi-modal data for effective biometric recognition. To enable the development, testing, and evaluation of these software systems, the BRIAR Testing and Evaluation Team has gone great lengths to build
and extend a one-of-a-kind dataset of images and video over the course of multiple data collection events. The BRIAR Government Collections 3 (BGC3) and 4 (BGC4) expand the BRIAR dataset \cite{briar-bgc2} to additional locations, more complex scenarios, and new sensors.

\begin{figure}
    \centering
    \includegraphics[width=0.95\linewidth]{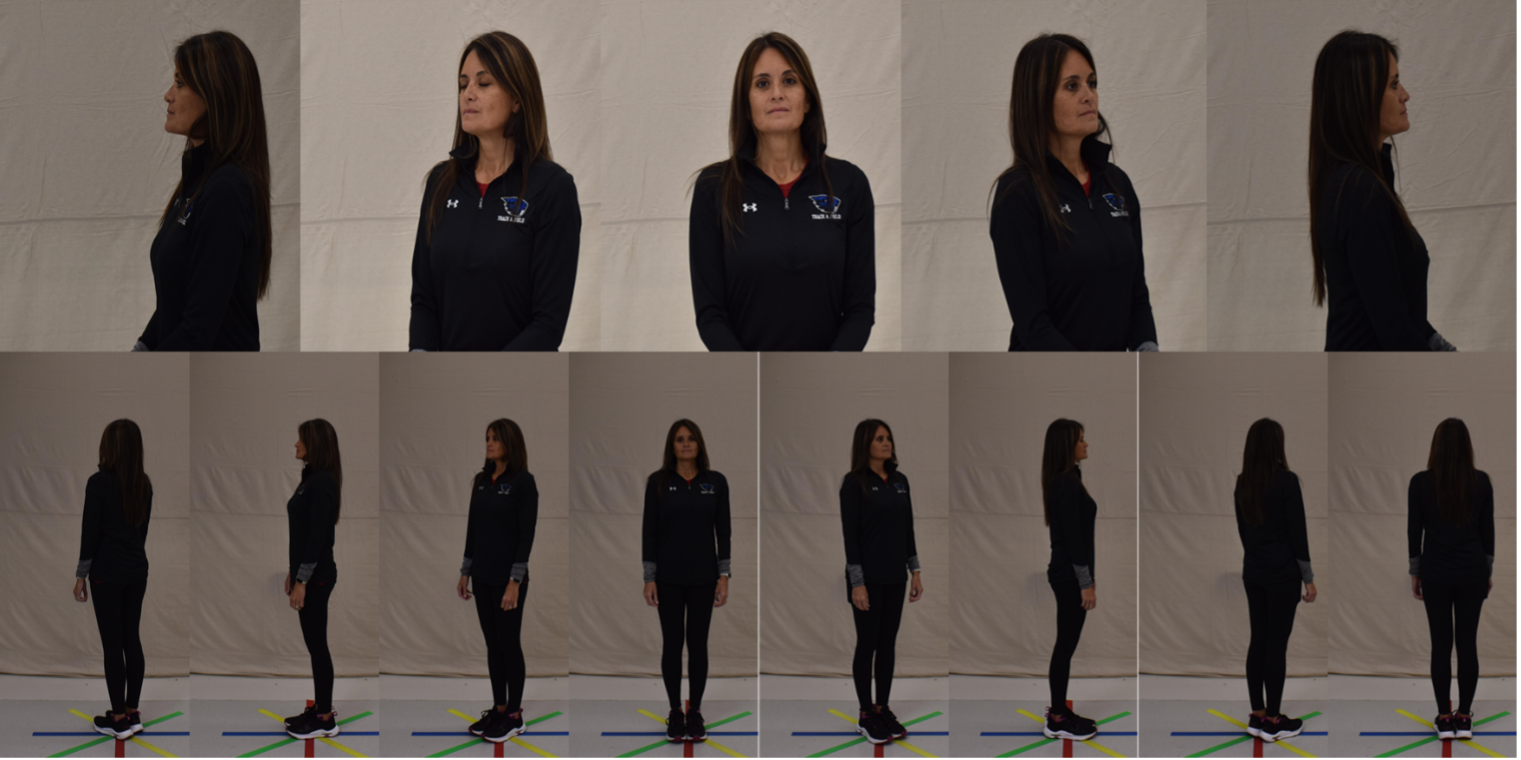}
    \caption{Sample images from the BRIAR dataset. Subjects in figures have consented to appearing in
publications.}
    \label{fig:image_set}
\end{figure}

\subsection{Contributions}
The introduction of the BRIAR dataset has been a monumental contribution to the biometrics community, and represents a major step forward for the computer vision community at large. It is the first dataset is of its kind, and has been a foundational benchmark in over 100 papers on biometrics and turbulence mitigation. Incorporating the BGC3 and BGC4 collections, the dataset consists of over 475,000 images and 3,450 hours of video of 1,760 subjects each in two sets of clothing, spanning three locations with varying climate and weather, captured using commercial- to military-grade and specialized cameras at ranges up to 1,000-m, at view angles up to 50\textdegree, and during both constrained and unconstrained imaging scenarios. Model development and testing is driven by continued expansion of the dataset and efforts to improve its quality by refining collection, curation, and annotation methods~\cite{briar_missions}. The addition of more diverse data, both in terms of the demographics pool of its enrolled subjects and the imaging conditions of the collection, will help to ensure that recognition models are equitable and robust~\cite{briar_insights}.

Accurate biometric identification systems have become a vital resource supporting security and safety initiatives. Beneficial applications are wide-ranging, from combating human trafficking and terrorism to supporting smart-city infrastructure and disaster response efforts. Even so, potential misuses of identification and recognition models as well as the data which shapes them are just as wide-ranging. Oversight and control of dataset access helps to prevent these potential misuses. For the BRIAR dataset, access requires approval from the US government sponsor and a data use agreement and review by the Institutional Review Board responsible for the dataset. Those interested in access to the BRIAR data should contact the authors, who will coordinate the request to the appropriate government contact.
The BRIAR dataset is already being utilized across a diverse range of research initiatives. 

The remainder of this paper is organized as follows: In 
Section~\ref{sec:background} we provide background information addressing the challenges and considerations in biometrics at long-range and high-altitude and a summary of related work in this topic area. Section~\ref{sec:methods} provides a detailed overview of the data collection methodology and highlights its unique elements. Section~\ref{sec: summary} includes statistical breakdowns of the most recent additions to the dataset. Section~\ref{sec:curation_methodology} details the challenges and methodology used in curating such a feature-rich dataset with emphasis on the curation pipeline, quality analysis, metadata construction, annotations, evaluation protocol design, and other associated activities. Finally, Section~\ref{sec:conclusions} provides discussions and concluding remarks including plans for future extensions to the ongoing program.

\section{BACKGROUND AND RELATED WORK}
\label{sec:background}

The widespread availability of datasets used for facial recognition has rapidly grown in recent years. However, the majority of these are focused on compliant biometric capturing and collected either by utilizing controlled acquisition or by scraping the internet for high quality images. The BRIAR dataset is unique in that it considers whole-body signatures (face, body, and gait) at long range and extreme pitch angles while participants perform a mixture of structured and open-ended activities under realistic operational conditions. 

BRIAR BGC1-4 is complemented by additional datasets collected under related efforts~\cite{accenture-mm1,mao2022single,davila2023mevid,HBRC-500,grover2024revealing,zhu2023gait}, many of which are openly available. There also exists a small body of work prior to the BRIAR program that considers the impact of long-range imaging to the performance of facial recognition models. Such works include the studies and datasets listed in Table \ref{tab:datasets}. Although these works demonstrate the need for further study, they do not provide the required experimental diversity to explore performance impacts related to clothing changes, multiple environmental conditions, sensor/optical configuration, subject demographics, or other factors that can only be teased out via a large and complex dataset.

\begin{table*}
\small
    \centering
    \begin{tabular}{|lcr|rr|c|}
        \hline
        \textbf{Dataset or Study Name} & \textbf{Citation} & \textbf{Year} & \textbf{Distance} & \textbf{Subjects} & \textbf{Notes} \\
        \hline
        \hline
        UTK-LRHM & \cite{utk-lrhm} & 2007 & 300m & 48 & Visible \\
        UMD Remote & \cite{umd-remote} & 2010 & 250m & 17 & Visible \\
        NFRAD & \cite{nfrad} & 2011 & 50m & 60 & NIR and Visible\\
        WVU FRAD NIR Mid-Range & \cite{multi-spectral} & 2012 & 120 m & 103 & NIR and Visible \\
        WVU FRAD DB3 Outdoor & \cite{multi-spectral} & 2012 & 400m & 16 & SWIR \\
        WVU FRAD DB2 Indoor & \cite{multi-spectral} & 2012 & 106m & 50  & SWIR \\
        UCCS Large Scale & \cite{open-set} & 2013 & 100m & 308 & Visible \\
        UMD LDHF & \cite{nighttime-standoff} & 2014 & 150m & 100 & NIR and Visible \\
        IJB–S Janus Surveillance & \cite{janus-s} & 2018 & 150m & 202 & Visible \\
        HBRC-500 & \cite{HBRC-500} & 2023 & 500m & 250 & Visible and LWIR\\
        Accenture-MM1 & \cite{accenture-mm1} & 2024 & 500m & 227 & Visible\\
        \textbf{BRIAR BGC 1-4} (this paper) & \cite{briar-bgc2} & 2024 & 1000m & 1173$^{**} $ & Visible, SWIR$^*$ MWIR$^*$, LWIR$^*$ \\
        \hline
    \end{tabular}
    \centering \caption{Comparison of the BRIAR dataset to similar large-scale, long-range, and/or multimodal studies. \\ $^*$ Nonvisible is currently held back for future research efforts. \\ $^{**}$ The dataset includes an additional 587 distractors with indoor images only.}
    \label{tab:datasets}
\end{table*}

\section{DATA COLLECTION METHODS}
\label{sec:methods}
The BGC3 and BGC4 collections build off the work and established procedures of previous collections and reuse much of the infrastructure and equipment of \cite{briar-bgc2}. These additional data collections for the BRIAR project incorporate new sensors, important updates to collection systems and software, new locations around the country, and new scenarios in order to produce useful data for the development and testing of robust recognition models.

The majority of the BRIAR dataset consists of videos and images collected at an indoor controlled location and an outdoor field setting, and features individual subjects. However, the BGC3 collection saw the introduction of group activities in the field, featuring multiple subjects. Additionally, the BGC4 collection participants were recorded during scenarios held in a mock-city setting, dubbed Hogan's Alley by the collection team, often featuring multiple subjects.

Continuous improvements are made to custom systems and software over the course of the program to address bugs and issues and to add additional tracking or collection capabilities given site- and collection-specific priorities and requirements.

\subsection{Privacy, Security, and Well-being}
\label{subsec: privacy}
Ethical research is an extremely important concern within the biometrics community; it is a concern that this work does not take lightly. All BRIAR dataset collection efforts are performed with a privacy-first and safety-first approach. Subject recruitment, informed consent, participation, and data handling procedures are approved by an Institutional Review Board (IRB), and the utmost care is taken to ensure that the data is collected and stored ethically and safely. Furthermore, the BRIAR dataset itself is de-identified, and does not associate any media with the actual identity of any of the participants. Instead, each subject is assigned a unique subject ID (e.g. G03045, G04237), which is used to label the data from the collection activities they participate in. Subjects may withdraw from the collection at any time, and their data cannot appear in publications without additional explicit consent.

\subsection{Controlled Collection}
\label{sec:controlled_collection}
The controlled scenarios were kept consistent with previous collections. The only exception was the separation of the ``random walk'' and ``cell phone'' activities \cite{briar-bgc2}, which were previously combined in a single recording.

Image data was collected at two stations, both of which had three Nikon DSLR cameras arranged on a 12-foot tall vertical stand. These cameras are triggered remotely to capture sets of images of the subject as they turn to face along several specified directions relative to the cameras, with one station collecting passport-style photos with neutral and smiling facial expressions, and the other collecting whole-body photos in a neutral standing position and with arms and legs extended in an x-shape pose (similar to TSA screening). Figure \ref{fig:image_set} provides some examples of the neutral expression and pose images. An application was developed as an extension to the existing BRIAR Human Subjects Testing Application (BHST App) to provide an interface to the cameras for capturing, tracking, and downloading images. The new system eliminated the extensive correction and manual checking needed due to the unreliability of the previous manual remote triggering, the independent internal clock of each DSLR camera, and human error.

\subsection{Field Collection}

\begin{figure}
    \centering
    \includegraphics[width=0.95\linewidth]{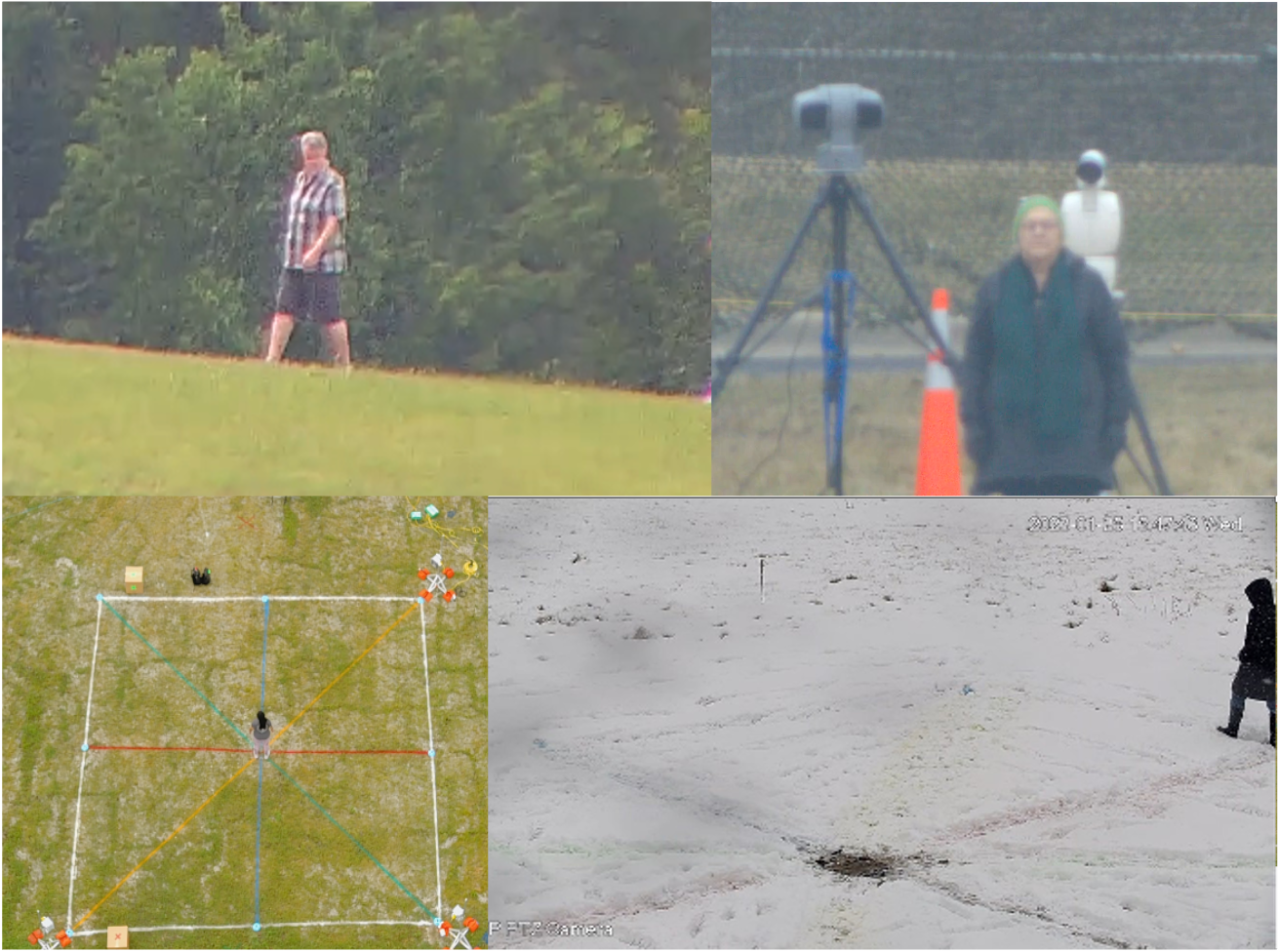}
    \caption{Sample frames of videos captured by various platforms during BGC3 and BGC4 (clockwise from the top-left): ground camera at 500m, rooftop camera at 720m, elevated close-range camera, UAV platform.}
    \label{fig:field_data}
\end{figure}

The field portion of the collection takes place in a 10x10-m square marked-off area, where subjects are instructed by a proctor to perform activities like standing facing along a series of colored lines or walking around randomly within the square. Subjects are recorded during these activities by commercial surveillance cameras, specialized long-range cameras, and unmanned aerial platforms focusing on the activity area from up to 500-m away in BGC3 and 720-m away in BGC4 (see Figure \ref{fig:field_data} for example images). Previously, subjects completed outdoor activities in both clothing sets. Starting with BGC3, subjects completed the outdoor field portion of the collection wearing only one clothing set in order to reduce the time required for participation, which could be up to four hours total.

\begin{figure}
    \centering
    \includegraphics[width=0.95\linewidth]{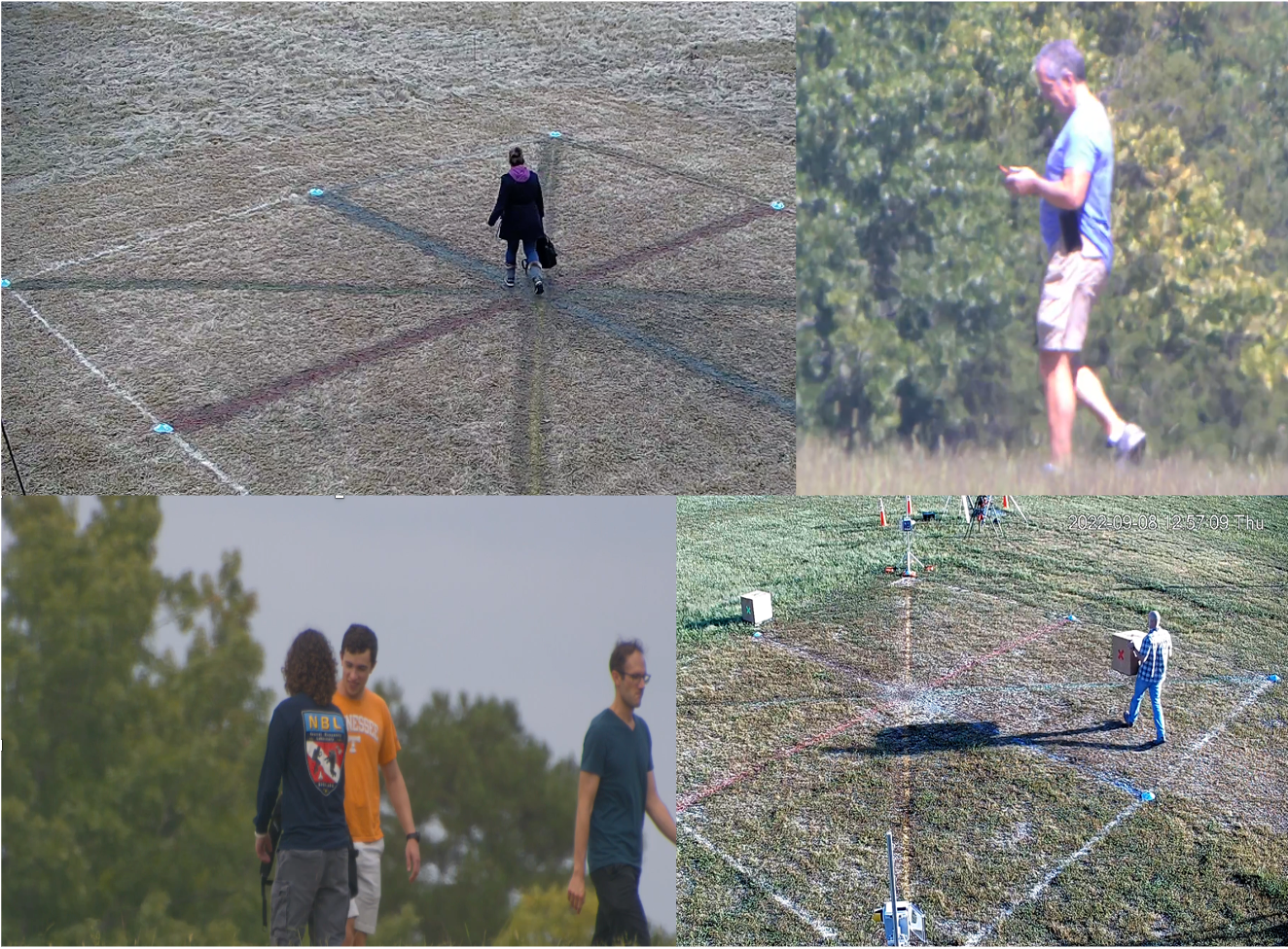}
    \caption{Frames of subjects participating in some of the new field scenarios (clockwise, starting from the top-left): ``backpack'', ``cell phone'', ``box stack'', and ``group backpack''.}
    \label{fig:field_activities}
\end{figure}

The field portion of the BRIAR collections saw more substantial changes following BGC2, including the introduction of new activities, examples of which can be seen in Figure \ref{fig:field_activities}:
\begin{itemize}
    \item \textbf{Cell Phone}: the subject is instructed to walk around randomly in the field area while pretending to text and receive a phone call.
    \item \textbf{Box Stack}: the subject is instructed to randomly place a cardboard box in the field area, stack another box on top of it, then return both to their original locations.
    \item \textbf{Backpack}: the subject is instructed to pick up and put on a weighted backpack in the center of the field area, walk around the area randomly, then return the backpack to the center.
    \item \textbf{Group Backpack}: Up to three subjects walk around randomly in the field area and pass a weighted backpack back and forth between them as they walk.
    \item \textbf{Pointing} (BGC4 only): the subject is instructed to look around and physically point to cameras and UAVs they can see.
\end{itemize}

\begin{figure}
    \centering
    \includegraphics[width=0.95\linewidth]{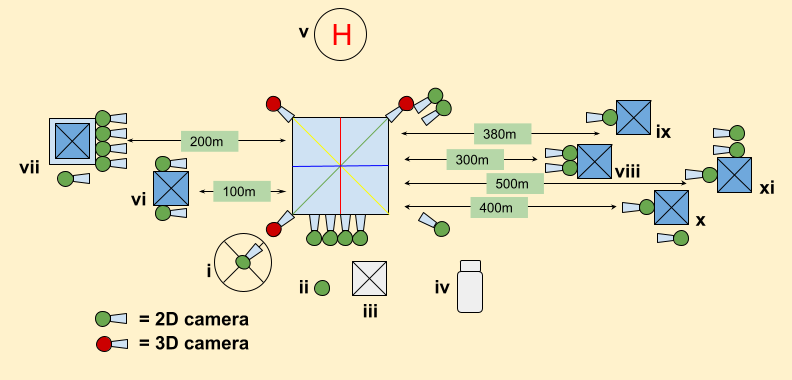}
    \caption{Overview of the BGC3 field collection layout. Labelled areas: (i) close-range mast-mounted cameras, (ii) fish-eye weather camera, (iii) outdoor collection tent, (iv) UAV control center, (v) UAV landing zone, (vi) 100-m range cameras, (vii) 200-m range cameras on scaffolding, (viii) 300-m range cameras, (ix) 380-m range cameras, (x) 400-m range cameras, (xi) 500-m range cameras}
    \label{fig:bgc3_field_layout}
\end{figure}

The BGC3 collection event was conducted over August and September, 2022, in Oak Ridge, Tennessee. The hosting location for BGC3 was the same as BGC1, and the geometry of the field deployment followed a similar template with a few significant differences (see Figure \ref{fig:bgc3_field_layout} for layout details). 

\begin{figure}
    \centering
    \includegraphics[width=0.95\linewidth]{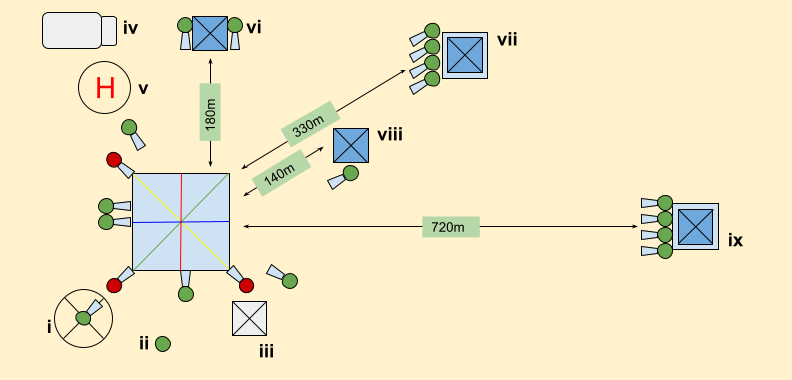}
    \caption{Overview of the BGC4 field collection layout. Labelled areas: (i) close-range mast-mounted cameras, (ii) fisheye weather camera, (iii) outdoor collection tent, (iv) UAV control center, (v) UAV landing zone, (vi) 180-m range cameras, (vii) 330-m range rooftop cameras, (viii) 140-m range camera, (ix) 720-m range rooftop cameras}
    \label{fig:bgc4_field_layout}
\end{figure}

BGC4 took place over the month of January, 2023, in a suburb of Chicago, Illinois. The low temperatures and snow not only presented interesting atmospheric imaging challenges, but also had a detrimental effect on the hardware and sensors used during the collection. Several sensors and pieces of equipment were damaged or malfunctioned, some had to be replaced, and some were not always accessible to be adjusted or fixed due to weather. View angles and locations were highly varied, with two stations located on rooftops, and most cameras shooting over a mix of asphalt and grass (see Figure \ref{fig:bgc4_field_layout} for layout details). The weather ran a gamut of fairly temperate conditions to frigid temperatures and significant snow accumulation towards the end of the collection.

\subsection{BGC4 Mock City: Hogan's Alley}

\begin{figure}
    \centering
    \includegraphics[width=0.95\linewidth]{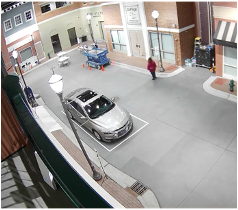}
    \caption{Example frame from a camera on the balcony of Hogan's Alley.}
    \label{fig:hogans_alley}
\end{figure}

A subset of the subjects were also recorded in scenarios taking place in an indoor street scene. The area was constructed to resemble a commercial/urban street setting with false storefronts, asphalt paving, sidewalks, streetlights, a fire hydrant, and other similar infrastructure. Balconies and windows on the upper story made it convenient to set up elevated camera views, as seen in Figure \ref{fig:hogans_alley}. In one corner, the collection team set up a mock market area with a table of snacks and smaller tables and chairs to sit at. A car was parked in the street for the subjects to interact with.  Individual subject activities in the mock city were not timestamped, instead, the proctor recorded only the entry and exit times of each subject. The BHST app was extended so that each subject's ID, clothing set number, and mock city entry and exit times could be recorded along with the rest of the normal collection data.

Participants were given fake money to use in the mock market and were instructed by proctoring staff to enter the street scene and complete activities posted on numbered signs throughout the area; however, the activities in the mock city were purposefully unstructured, and participants were not required to complete any of them. Several participants could be active at one time. Generally, subjects would enter the mock city after they had completed the field and controlled portions of the collection, and once several subjects had completed most of the activities, they would exit as a group. 

\begin{figure}
    \centering
    \includegraphics[width=0.95\linewidth]{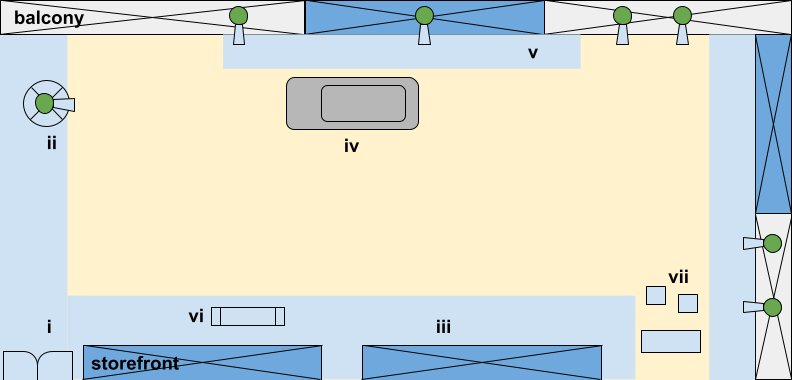}
    \caption{Overview of the BGC4 mock city layout: a simulated street scene. Labelled areas: (i) subject entrance/exit, (ii) lamppost mounted camera, (iii) bank storefront, (iv) parked car, (v) pharmacy storefront, (vi) bench, (vii) mock market/cafe setup}
    \label{fig:bgc4_hogans_layout}
\end{figure}

Figure \ref{fig:bgc4_hogans_layout} shows the locations of the suggested scenarios, where the subjects were instructed to e.g. sit on the bench for 30 seconds, remove and replace a box from the backseat of the car, or use fake money to purchase snacks from the market.

The Hogan's alley scenarios and data collection were intended to produce more naturalistic data for research and development with several subjects coming in and out of camera views, performing unstructured, everyday actions in videos that are not strictly curated. 

\section{DATASET SUMMARY}
\label{sec: summary}
The figures below summarize statistics for the BGC3 and BGC4 data sets. The BGC3 data set consists of over 45,000 images and 38,000 videos. The BGC4 data set consists of over 80,000 images and 45,000 videos. This data is further split into the BRIAR Research Set (BRS) and BRIAR Test Set (BTS) intended for model training and testing respectively. The complete curated data of each subject is assigned to one of these subsets with the intention of keeping the distribution of subject demographics consistent between BRS and BTS datasets.

\begin{figure}[t!]
    \centering
    \includegraphics[width=0.95\linewidth]{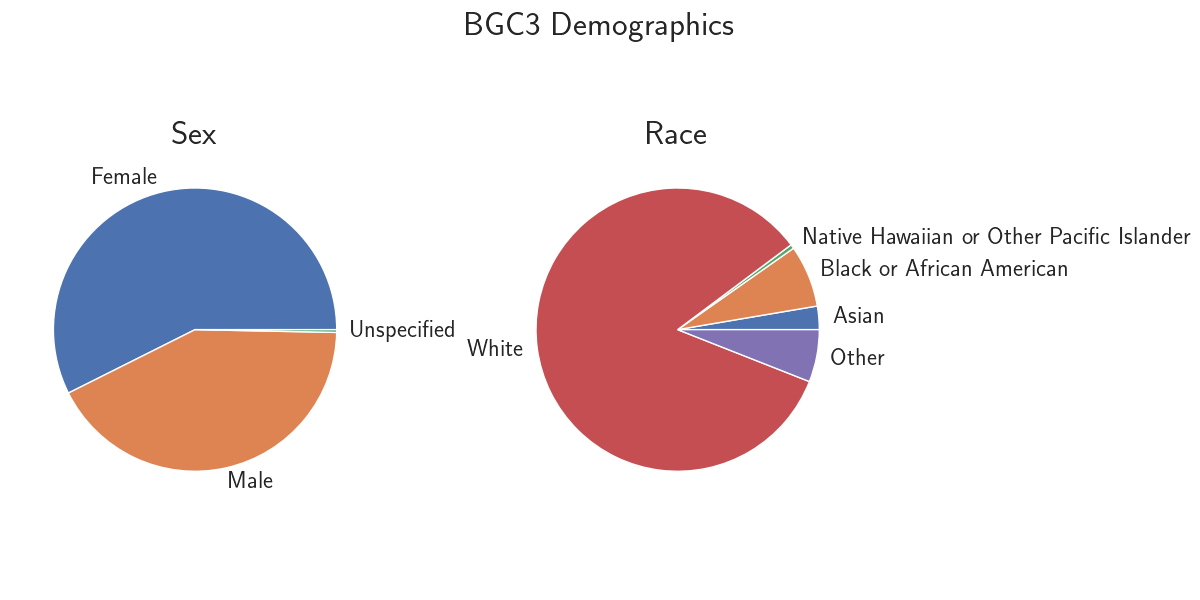} \\
    \includegraphics[width=0.95\linewidth]{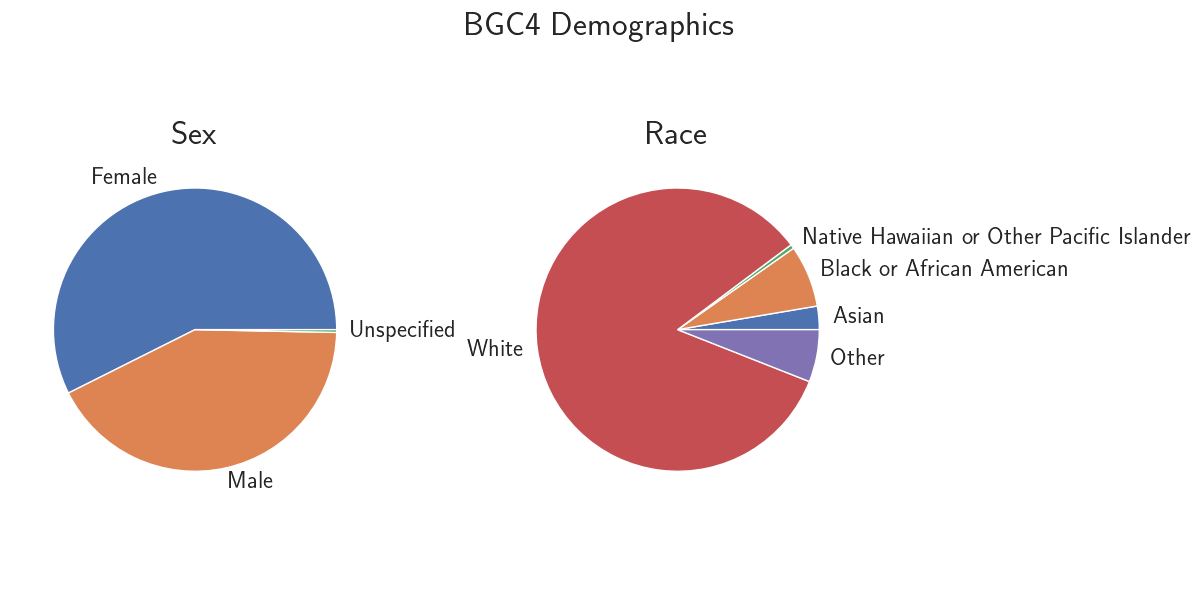}
    \caption{Overview of sex and ethnicity statistics for BGC3 (top) and BGC4 (bottom) subjects}
    \label{fig:BGC34-ethnicity-demographics}
\end{figure}

\begin{figure}[t!]
    \centering
    \includegraphics[width=0.95\linewidth]{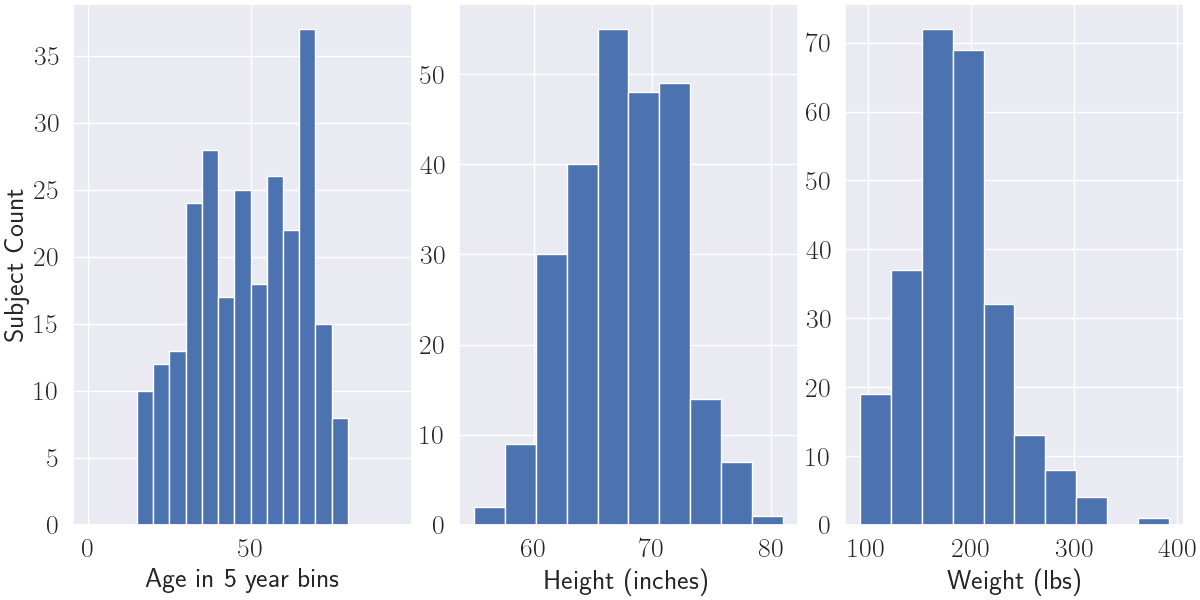} \\
    \includegraphics[width=0.95\linewidth]{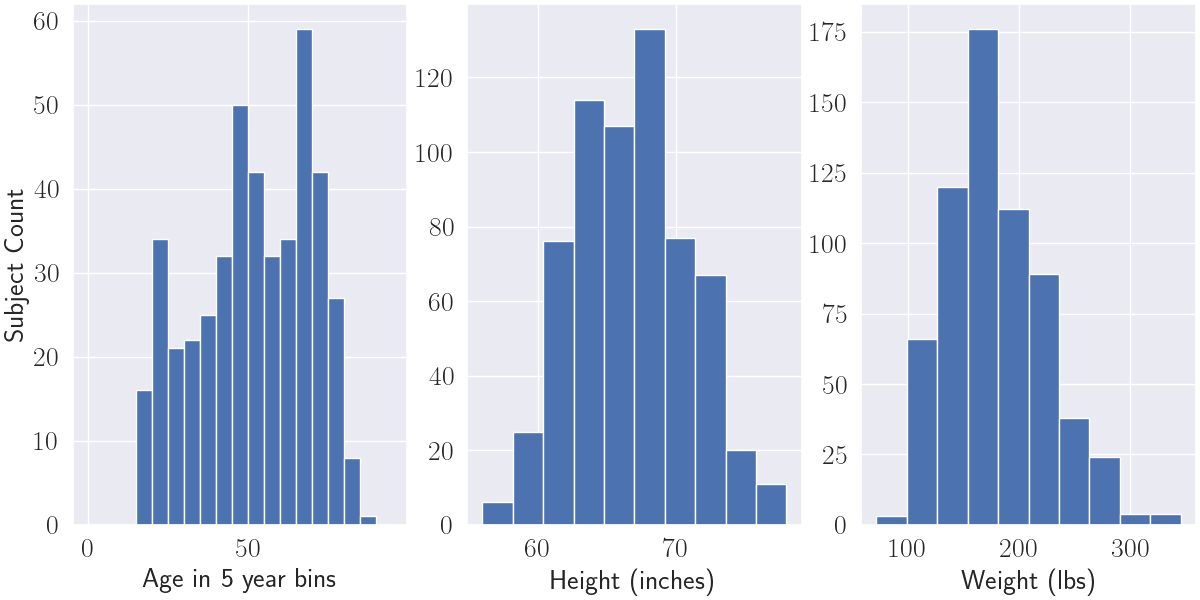}
    \caption{Overview of age, height, weight statistics for BGC3 (top) and BGC4 (bottom) subjects}
    \label{fig:BGC34-age-demographics}
\end{figure}

Figures \ref{fig:BGC34-ethnicity-demographics} and \ref{fig:BGC34-age-demographics} show the distribution of subject sex, race, age, height and weight of BGC3 and BGC4 subjects, respectively. Age, height, and weight have similar, approximately normal distributions across the two data sets. The two data sets also have nearly identical makeups of subject sexes, with specified sexes being slightly more female than male. Both BGC3 and BGC4 subjects overwhelmingly report ``white'' as their race.

\begin{figure}[h!]
    \centering
    \includegraphics[width=0.99\linewidth]{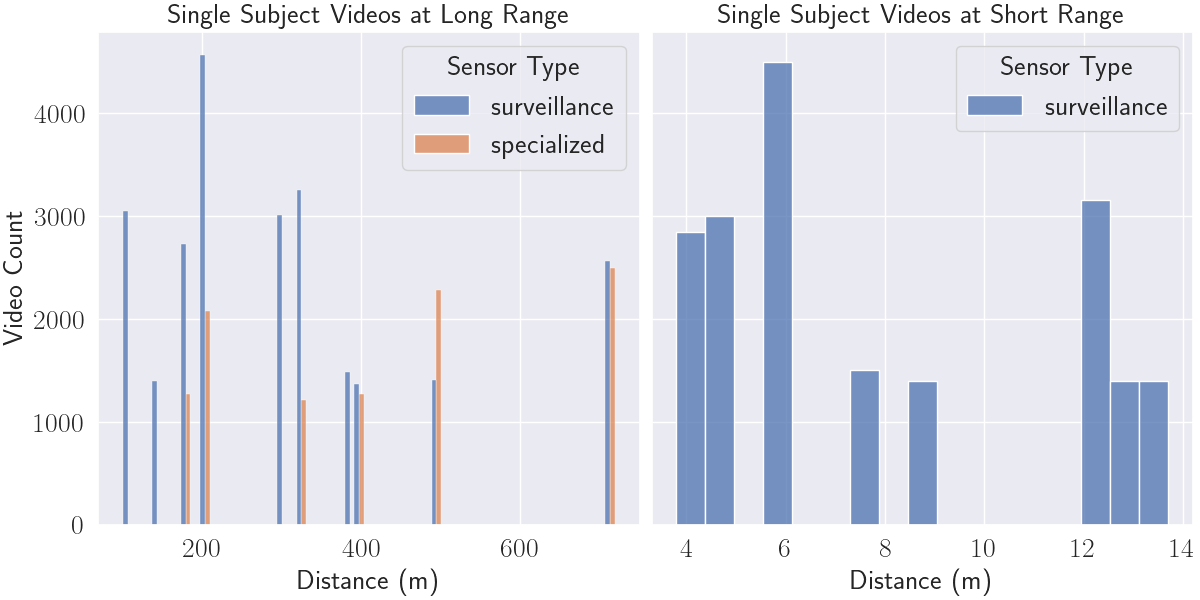}\\
    
    \vspace{.4cm}
    
    \includegraphics[width=0.99\linewidth]{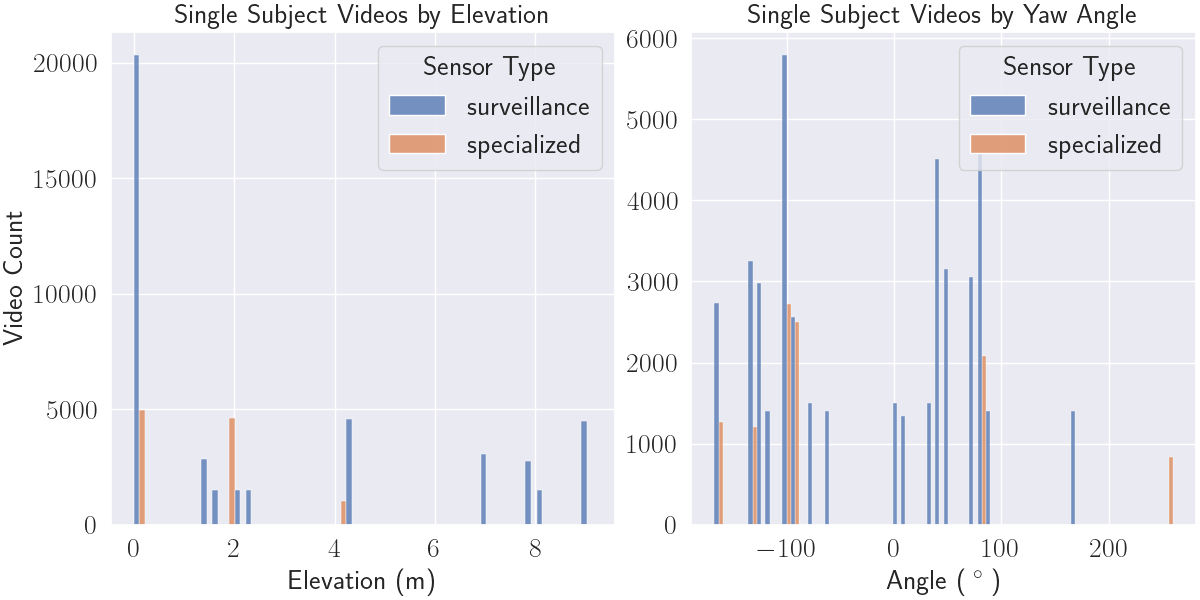}
    \caption{Number of videos captured of a single subject per distance, elevation, and yaw angle relative to subject location}
    \label{fig:single-video-distributions}
\end{figure}

\begin{figure}[t!]
    \centering
    \includegraphics[width=0.99\linewidth]{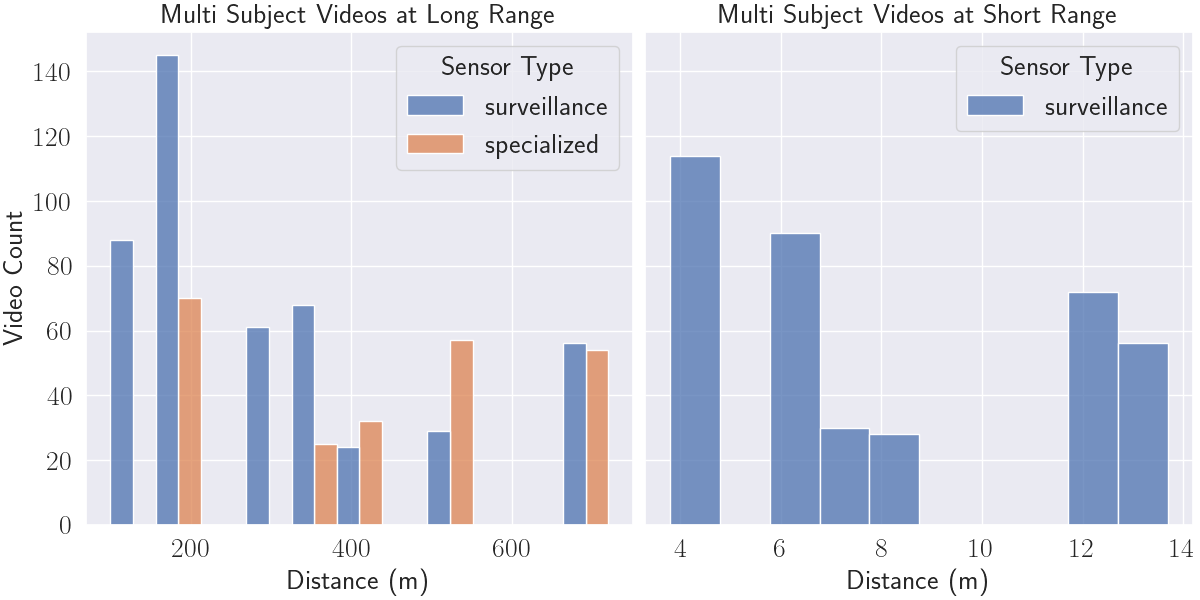}\\
    
    \vspace{.4cm}
    
    \includegraphics[width=0.99\linewidth]{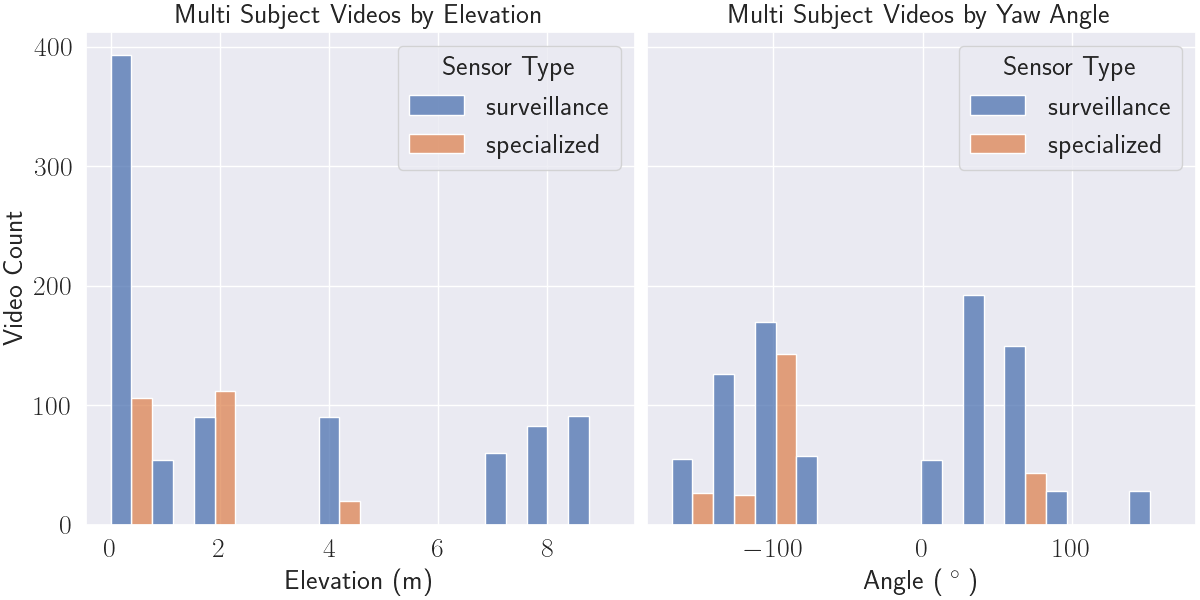}
    \caption{Number of multi subject videos captured per distance, elevation and yaw angle relative to subject group location}
    \label{fig:group-video-distributions}
\end{figure}

Figure \ref{fig:single-video-distributions} shows the distribution of the number of single  BGC3/4 subject videos among distance, elevation, and yaw angle relative to the subject. Figure \ref{fig:group-video-distributions} shows the distribution of the number of BGC3/4 group videos among distance, elevation, and yaw angle relative to the group.

\section{DATA CURATION METHODOLOGY}
\label{sec:curation_methodology}
The following section details the specific strategies and technologies used in the annotation and curation process of the BRIAR datasets. This section additionally discusses upgrades and changes to procedures used to curate the previous BGC1/2 datasets~\cite{briar-bgc2}.
\begin{table*}[]
\footnotesize
\centering
\begin{tabular}{|l|l|l|l|l|}
\hline
                                                                                  & Surviellance & Specialized & UAVs & Images \\ \hline
\textbf{1. Data Cleaning and Compilation (Section~\ref{subsubsec:cleaning})}                                                  &              &             &      &        \\ \hline
a. Automated Timestamp validation via internal clock                       & X            & X           & X    & X      \\ \hline
b. Manually Update sensor position metadata                                             & X            & X           &      & X      \\ \hline
c. Re-annotate corrupt timestamps                                &              &             & X    &        \\ \hline
d. Match recordings to BHST subject trigger timestamps                                           &              & X           &      &        \\ \hline
\textbf{2. Video and Image Extraction (Section~\ref{subsubsec: extraction})}                                                     &              &             &      &        \\ \hline
a. Cut videos clips unique to each activity/subject/station/clothing set        & X            & X           & X    &        \\ \hline
b. Convert raw NEF images to JPEG                                              &              &             &      & X      \\ \hline
\textbf{3. XML Data Generation (Section~\ref{subsubsec: metadata})}                                                            &              &             &      &        \\ \hline
a. Link subject metadata* to images and video         & X            & X           & X    & X      \\ \hline
b. Link Field Measurements** to images and video          & X            & X           & X    &        \\ \hline
c. Link subject metadata with individual activities                      & X            & X           & X    &        \\ \hline
d. Record all linked information in associated XML file                                     & X            & X           & X    & X      \\ \hline
\textbf{4. Quality Assurance and Finalization (Section~\ref{subsubsec: qa})}                                            &              &             &      &        \\ \hline
a. Validate all videos/images for first and last subject of collection each day                & X            & X           & X    & X      \\ \hline
b. Sanity-check metadata values recorded for first and last subjects of each day & X            & X           & X    & X      \\ \hline
c. Validate XMLs against XSD schema                                                  & X            & X           & X    & X      \\ \hline
d. Partition videos and images into BRS/BTS splits                                         & X            & X           & X    & X      \\ \hline
\end{tabular}
\caption{Overview of data curation pipeline separated into relevant tasks by data type. Each row in the table denotes a sequential step in the annotation and curation process. 
*subject metadata includes demographics, activities, and camera specifications
**field measurements include activities, weather, CN2, camera measurements}
\label{table:data-curation-pipeline}
\end{table*}


\subsection{Data Curation Pipeline}
\label{subsec:pipeline}
The collected data was curated such that there is no more than one video or image for each unique subject, activity, clothing set, and sensor. For each curated video a corresponding XML file was generated that describes the activity details, subject demographics, camera specifications, camera measurements, and weather/atmospheric conditions. After curation, a set of quality assurance steps were taken to improve the overall quality of the final dataset. Each category of sensor warrants a slightly different curation process and so each has its own curation pipeline. Similarities and differences between these pipelines are shown in Table \ref{table:data-curation-pipeline}.

\subsubsection{Data Cleaning and Compilation}
\label{subsubsec:cleaning}
Before videos and images were extracted, the collection's raw data had to be organized and checked for inconsistencies. It was expected that there would be a few issues with the metadata detailing which subjects performed certain activities: ferrying subjects between multiple concurrently running collection locations and dealing with de-identified subject ID numbers is bound to produce mistakes. Initial validation tests were run to identify and fix these mistakes. These scripts ensured that timestamp data did not show a subject appearing in two locations or activities at once, and that there were no overlapping time records of different subjects performing an activity at the same location. For the timestamp collisions that were detected, video from the locations and times in question were manually checked and cross referenced with known imagery of subjects to correct the records. As a second initial validation test, a few videos from each day of collection were randomly selected and spot-checked at specific timestamps that could be verified against a known subject activity.

Once these to steps were complete, the surveillance, specialized, and still image sensors were associated with manual measurements taken during the collections. These measurements include the distance to subject, sensor height, sensor yaw in relation to the subject, and sensor pitch angle. In addition to these measurements, all sensor specifications are recorded in a csv file detailing the manufacturer, model, minimum / maximum focal lengths, and capture spectrum.

The specialized cameras and UAV platforms required additional data processing due to their non-standard formats and configurations. The long-range specialized R\&D cameras capture pre-cut raw and compressed video recordings via timing triggers broadcast over UDP by the BHST app. The format and set of metadata tags included in the videos recorded by the UAV platforms were often unique to that platform, so several tools in combination were required to extract their timestamp metadata. Additionally, because they could not be connected to the GPS-synchronized Network Time Protocol (NTP) servers~\cite{andrianto2018performance} that set the internal clocks of the ground sensors and recording systems, the UAV time metadata was manually aligned by visually matching to video from a synchronized camera. 

\subsubsection{Video and Image Extraction}
\label{subsubsec: extraction}
 Referencing the timestamps recorded at the start and end of each subject activity, each video was cut into segments using FFMPEG~\cite{FFmpeg}. The videos were cut such that each curated segment is specific to a subject, clothing set, and activity. All videos with a compatible codec were converted to an .mp4 container and non-compatible file formats were preserved. Raw images were captured and stored in a .NEF container. These raw files are provided in the curated dataset as well as an accompanying .jpg file that was extracted directly from the raw file.

\subsubsection{XML Metadata Generation}
\label{subsubsec: metadata}
In order to provide context for all of the activities that were recorded during the collection event, a corresponding metadata file was generated for each curated video. All of the information used to populate this metadata file was loaded at this stage in the pipeline. The final metadata file contains information regarding the subject demographics, the weather conditions, and scintillometer (CN2) readings at the time of the activity, as well as the sensor details described above. Weather and atmospheric details in the metadata file include: temperature, wind chill, heat index, relative humidity, wind speed, wind direction, barometric pressure, and solar loading. An XSD schema is used to ensure XMLs follow a standard format, and automated tests are run against the XML data to ensure realistic values are recorded. The XSD schema is included along with dataset documentation.

\subsubsection{Quality Assurance and Finalization}
\label{subsubsec: qa}
To verify that the data was curated correctly, manual annotators checked every video and image from the first and last subject of each day during the collection event and validated their timestamps, content, and accompanying metadata. Once these checks were completed, the data was partitioned into BRIAR Research Set (BRS) and BRIAR Test Set (BTS) data, meant for training and testing, respectively. The BRS and BTS sets are subject-disjoint and balanced with respect to subject demographics.

\subsubsection{Group Activity Curation}

Aside from the standard single subject activities there were also activities which involved multiple subjects participating in one video. The curation process for these activities was very similar to the standard curation pipeline with a few key differences. In the single subject activity videos the subject ID was used in the directory structure of the final curated video path. Instead of listing multiple IDs, the group scenario videos were placed into a separate directory and labeled with a unique group scenario identifier. The XML metadata files contain multiple subject demographic sections, one for each subject in the recorded activity. Because of their unstructured nature, some of the longer BGC4 mock city scenarios were evenly split into smaller segments such that no curated video exceeds 20 minutes in duration. Like the field group scenario videos, each of the segments was assigned a unique group scenario identifier, with the majority of videos not requiring a split. 

\subsection{Annotations}

Automated annotations were generated using a chain of open-source and pre-trained models. Whole-Body (WB) detection was done with YOLOv5 and a fine-tuned version was used on long-range and aerial videos \cite{yolov5}. 3D human mesh reconstruction with Meshtransformer and 2D keypoint estimation with DARK was performed on the WB detection results \cite{Zhang_2020_CVPR, DBLP:journals/corr/abs-2012-09760}. Re-ID with DG-Net++ was then performed on the pose results to determine whether or not a WB detection was the intended subject~\cite{zou2020joint}. The pose information helped narrow the gallery to reference images at a similar yaw angle to the detections. BoT-SORT was used for track generation, which leveraged the Re-ID results for better track consistency \cite{aharon2022botsort}. Finally, various post-processing steps were performed, such as estimating the head bounding box from the 3D mesh and 2D keypoints and removing spurious tracks.

Manual annotations were performed on specific video frames to either verify or correct the automatic annotations. The main task given to manual annotators was to verify that the correct subject was associated to a given track, because the main goal of the project is identification. To save on cost given the size of the dataset, only the first and last frame of a track were used for verification instead of every frame of a track. Other tasks involve manual annotation of frame ranges with unexpected missing whole-body detections, which is more common in low image quality conditions. The resulting sparse manual annotations were then merged with automatic annotations to produce an XML with metadata and annotations for each video in the dataset. Additionally, any non-subject persons that were visible, such as data collection proctors, were censored by insertion of a black rectangle in the video to satisfy IRB protocols.

\subsection{Evaluation Protocol Design}
The evaluation protocol design incorporates all BTS data collected to date, reflecting the growing complexity of the BRIAR program. The BTS set is partitioned into probe and gallery sets with probe sets further categorized into two major types: FaceIncluded and FaceRestricted. The FaceIncluded probe set contains data where faces are visible and have a head height of at least 20 pixels, ensuring that each subject in every probe has at least one detectable face. In the FaceRestricted probe set, all faces are either occluded, have low resolution (less than 20 pixels in head height), or are not present. Pose estimation was the key parameter used to group these categories.
The probe set in the evaluation protocol is composed of data from long-range cameras, close-range cameras with elevations up to 50° and UAVs. Each probe consists of 5- to 15-second video clips that are extracted from the captured activities in the field.  Biometric algorithms are tasked with identifying these probes against the people found in the gallery.
The evaluation protocol design for the BRIAR dataset utilizes two types of galleries, \textbf{simple} and \textbf{blended}.
\begin{itemize}
    \item \textbf{Simple}: This collection features various body and face images captured from different perspectives. It includes walking video sequences from various angles to support gait recognition.  It is intended to represent an ideal enrollment for whole body and face recognition algorithms.
    \item \textbf{Blended}: While the Simple gallery serves as a baseline throughout the phases in the BRIAR Program, the Blended gallery was introduced to simulate more realistic conditions. It is named ``blended'' because 60\%\ of the subjects utilize a full gallery, identical to the Simple gallery format. However, 40\%\ of the subjects have fewer images, incorporating a smaller number of walking sequences for both indoor and outdoor settings. For 20\%\ of subjects, media are chosen from mugshot photos and indirect angles from close-range ground and elevated surveillance camera feeds using both field and controlled collection data. The remaining 20\%\ of subjects’ media are chosen from walking sequences that include only views directly facing the camera. This simulates realistic operational data found in law enforcement databases including mugshot like images and video from security choke points.
\end{itemize}

\section{CONCLUSIONS AND FUTURE WORKS}
 
\label{sec:conclusions}
The BRIAR Program is continuing to refine and expand its dataset, incorporating additional locations, subject scenarios, and complexity. Currently, BGC5 and BGC6 remain to be fully curated and released. These releases, BGC3 and BGC4, have placed a greater emphasis on data collection with an expanded range, locations, and conditions: particularly winter weather and clothing. Additional group scenarios and the mock city data provide new challenges relating to occlusion, detection, and tracking. These updates enable researchers to develop more generalizable models that can better handle a wider range of conditions.

Future work will focus on enhancing data quality through improved curation and annotation processes. The goal is to develop new methods that can measurably improve both the BRS and BTS datasets.

To date, the dataset comprises 1,173 full subjects and 587 distractors, encompassing over 475,000 images and 3,450 hours of video. This extensive resource is designed to support research in biometric identification at long-range and from elevated positions, ultimately contributing to critical security and intelligence requirements.

\section{ACKNOWLEDGMENTS}

This research is based upon work supported by the Office of the Director of National Intelligence (ODNI), Intelligence Advanced Research Projects Activity (IARPA), via D20202007300010. The views and conclusions contained herein are those of the authors and should not be interpreted as necessarily representing the official policies, either expressed or implied, of ODNI, IARPA, or the U.S. Government. The U.S. Government is authorized to reproduce and distribute reprints for governmental purposes notwithstanding any copyright annotation therein.

This research used resources from the Knowledge Discovery Infrastructure at the Oak Ridge National Laboratory, which is supported by the Office of Science of the U.S. Department of Energy under Contract No. DE-AC05-00OR22725.

Notice:  This manuscript has been authored by UT-Battelle, LLC, under contract DE-AC05-00OR22725 with the US Department of Energy (DOE). The US government retains and the publisher, by accepting the article for publication, acknowledges that the US government retains a nonexclusive, paid-up, irrevocable, worldwide license to publish or reproduce the published form of this manuscript, or allow others to do so, for US government purposes. DOE will provide public access to these results of federally sponsored research in accordance with the DOE Public Access Plan
(http://energy.gov/downloads/doe-public-access-plan).

The authors of this paper also wish to acknowledge the contributions of:

Knowledge Discovery Infrastructure (KDI) Staff: Dallas Sacca and Ryan Tipton.

Proctors and drivers: Raymond Borges-Hink, Nancy Engle, Dale Hensley, Nikki Jones, Michael Jones, Marylin Langston, Matt Love, Amanda Mottern, Gio Pascascio, Linda Paschal, Christina Peshoff, Donna Pierce, Ryan Styles, and Lauren Torkelson.

UAS Pilots: Joe Baldwin, Kase Clapp, Dakota Haldeman, Andrew Harter, Amanda Killingsworth, Matt Larson, Genevieve Martin, Aaron O'Toole, Jason Richards, and Brad Stinson.

Health and Safety Support Staff: Margaret Smith and Miranda Liner.


\section*{ETHICAL IMPACT STATEMENT}
As discussed in subsection \ref{subsec: privacy}, safe and ethical collection and data management practices are the first priority for the operation of the BRIAR team. 

The BRIAR program is reviewed and approved by the Central Department of Energy Institutional Review Board. The study number is 00000094. The IRB package included a protocol, consent, scripts, checklists, and surveys. A modification was submitted for any change to the research, procedures, or team members.
 
Each participant signed a consent form before participating. They had the opportunity to ask questions of the designated member of team responsible for the consent process. The participant was notified of how their data would be shared and were offered the opportunity to allow their photos, videos, and likeness to be used publicly by checking a box on the consent form. This was optional and the participant was notified it was their decision.  
 
The study posed minimal risk to the participants. The primary risk to subjects participating in BRIAR collections is the risk to privacy from disclosure of personally identifiable information (PII), i.e. a subject's appearance and likeness. To mitigate this risk, the BRIAR datasets are stored on secure, access-controlled computers in accordance with federal requirements for the protection of PII. Any sharing of the data requires approval and a signed data use agreement (DUA).

Risks to the physical safety of participants in the study were addressed as comprehensively as possible. Subjects included in outdoor collections were offered insect repellent, sunscreen, water, snacks, shade, and bathroom facilities to mitigate any risk to being outdoors. Onsite first aid care was provided at no cost to participants. All BRIAR team members were required to be certified in first aid and CPR. Participants were always accompanied by a BRIAR team member to ensure they would not enter into areas not associated with the study and in which the participant could encounter more significant risks.
 
Unmanned aerial systems (UAS) were operated in support of this project.  A mix of fixed-wing, tethered rotary-wing, and untethered rotary-wing aircraft were used.  All aircraft were operated by certified FAA Part 107 licensed flight crews in accordance with applicable regulations and documented safety protocols. Those safety protocols dictate minimum and maximum altitudes, maintenance and inspection requirements, flight procedures, crew requirements, weather restrictions, and other safety requirements in a variety of operational conditions.
 
Participants were compensated based on completion of indoor and outdoor activities. They received a gift card at the end of their participation. This amount was determined based on the time and effort given by the participant and was approved by the IRB to be appropriate to prevent monetary coercion.
 
The study did include participants associated with the employer of the BRIAR team. This is considered a vulnerable population by the IRB. The research team implemented additional measures to protect against coercion, including ensuring that individuals recruited did not report to any BRIAR study team member. The collections have included some university locations; in each case, the university was notified of the study for review.

Access to the dataset is managed by the US Government sponsor of the BRIAR program, and requires IRB review. These measures help to ensure that data use is ethical and follows US laws and IRB regulations for the protection of civil liberties.


{\small
\bibliographystyle{ieee}
\bibliography{egbib}

\begin{thebibliography}{10}\itemsep=-1pt

\bibitem{FFmpeg}
{F}{F}mpeg --- ffmpeg.org.
\newblock \url{https://ffmpeg.org/}.

\bibitem{aharon2022botsort}
N.~Aharon, R.~Orfaig, and B.-Z. Bobrovsky.
\newblock Bot-sort: Robust associations multi-pedestrian tracking, 2022.

\bibitem{andrianto2018performance}
H.~Andrianto, Y.~Susanthi, and D.~Suryadi.
\newblock Performance evaluation of low-cost gps time server based on ntp.
\newblock {\em TELKOMNIKA (Telecommunication Computing Electronics and
  Control)}, 16(6):2528--2535, 2018.

\bibitem{briar_missions}
D.~Aykac, J.~Brogan, N.~Barber, R.~Shivers, B.~Zhang, D.~Sacca, R.~Tipton,
  G.~Jager, A.~Garret, M.~Love, J.~Goddard, D.~Cornett, and D.~S. Bolme.
\newblock Long-range biometric identification in real world scenarios: A
  comprehensive evaluation framework based on missions.
\newblock In {\em 2024 IEEE International Joint Conference on Biometrics
  (IJCB)}, pages 1--9, 2024.

\bibitem{briar_insights}
D.~S. Bolme, D.~Aykac, R.~Shivers, J.~Brogan, N.~Barber, B.~Zhang, L.~Davies,
  and D.~Cornett.
\newblock From data to insights: A covariate analysis of the iarpa briar
  dataset for multimodal biometric recognition algorithms at altitude and
  range.
\newblock In {\em 2024 IEEE International Joint Conference on Biometrics
  (IJCB)}, pages 1--9, 2024.

\bibitem{multi-spectral}
T.~Bourlai and B.~Cukic.
\newblock Multi-spectral face recognition: Identification of people in
  difficult environments.
\newblock In {\em 2012 IEEE International Conference on Intelligence and
  Security Informatics}, pages 196--201, 2012.

\bibitem{briar-bgc2}
D.~Cornett, J.~Brogan, N.~Barber, D.~Aykac, S.~Baird, N.~Burchfield, C.~Dukes,
  A.~Duncan, R.~Ferrell, J.~Goddard, G.~Jager, M.~Larson, B.~Murphy,
  C.~Johnson, I.~Shelley, N.~Srinivas, B.~Stockwell, L.~Thompson, M.~Yohe,
  R.~Zhang, S.~Dolvin, H.~J. Santos-Villalobos, and D.~S. Bolme.
\newblock Expanding accurate person recognition to new altitudes and ranges:
  The briar dataset.
\newblock In {\em 2023 IEEE/CVF Winter Conference on Applications of Computer
  Vision Workshops (WACVW)}, pages 593--602, 2023.

\bibitem{davila2023mevid}
D.~Davila, D.~Du, B.~Lewis, C.~Funk, J.~Van~Pelt, R.~Collins, K.~Corona,
  M.~Brown, S.~McCloskey, A.~Hoogs, et~al.
\newblock Mevid: Multi-view extended videos with identities for video person
  re-identification.
\newblock In {\em Proceedings of the IEEE/CVF Winter Conference on Applications
  of Computer Vision}, pages 1634--1643, 2023.

\bibitem{HBRC-500}
C.~N. Fondje, K.~Nikhal, J.~B. Peace, R.~Karl, M.~W. Lee, P.~Berkowitz,
  K.~Gramzinski, B.~Kennedy, N.~Uzuegbunam, V.~Ou, T.~Barret, O.~Arend,
  W.~Ming, S.~Semenova, and B.~Riggan.
\newblock Hbrc-500: A long range recognition benchmark dataset using face and
  whole-body imagery.
\newblock In {\em 2023 IEEE International Joint Conference on Biometrics
  (IJCB)}, pages 1--11, 2023.

\bibitem{grover2024revealing}
S.~Grover, V.~Vineet, and Y.~Rawat.
\newblock Revealing the unseen: Benchmarking video action recognition under
  occlusion.
\newblock {\em Advances in Neural Information Processing Systems}, 36, 2024.

\bibitem{yolov5}
G.~Jocher, {Ayush Chaurasia}, A.~Stoken, J.~Borovec, {NanoCode012}, {Yonghye
  Kwon}, {TaoXie}, {Kalen Michael}, {Jiacong Fang}, {Imyhxy}, {, Lorna},
  C.~Wong, {(Zeng Yifu)}, {Abhiram V}, D.~Montes, {Zhiqiang Wang}, C.~Fati,
  {Jebastin Nadar}, {Laughing}, {UnglvKitDe}, {Tkianai}, {YxNONG}, P.~Skalski,
  A.~Hogan, M.~Strobel, M.~Jain, L.~Mammana, and {Xylieong}.
\newblock ultralytics/yolov5: v6.2 - yolov5 classification models, apple m1,
  reproducibility, clearml and deci.ai integrations, 2022.

\bibitem{janus-s}
N.~D. Kalka, B.~Maze, J.~A. Duncan, K.~O’Connor, S.~Elliott, K.~Hebert,
  J.~Bryan, and A.~K. Jain.
\newblock Ijb–s: Iarpa janus surveillance video benchmark.
\newblock In {\em 2018 IEEE 9th International Conference on Biometrics Theory,
  Applications and Systems (BTAS)}, pages 1--9, 2018.

\bibitem{nighttime-standoff}
D.~Kang, H.~Han, A.~K. Jain, and S.-W. Lee.
\newblock Nighttime face recognition at large standoff: Cross-distance and
  cross-spectral matching.
\newblock {\em Pattern Recognition}, 47(12):3750--3766, 2014.

\bibitem{DBLP:journals/corr/abs-2012-09760}
K.~Lin, L.~Wang, and Z.~Liu.
\newblock End-to-end human pose and mesh reconstruction with transformers.
\newblock {\em CoRR}, abs/2012.09760, 2020.

\bibitem{nfrad}
H.~Maeng, H.-C. Choi, U.~Park, S.-W. Lee, and A.~K. Jain.
\newblock Nfrad: Near-infrared face recognition at a distance.
\newblock In {\em 2011 International Joint Conference on Biometrics (IJCB)},
  pages 1--7, 2011.

\bibitem{mao2022single}
Z.~Mao, A.~Jaiswal, Z.~Wang, and S.~H. Chan.
\newblock Single frame atmospheric turbulence mitigation: A benchmark study and
  a new physics-inspired transformer model.
\newblock In {\em European Conference on Computer Vision}, pages 430--446.
  Springer, 2022.

\bibitem{umd-remote}
J.~Ni and R.~Chellappa.
\newblock Evaluation of state-of-the-art algorithms for remote face
  recognition.
\newblock In {\em 2010 IEEE International Conference on Image Processing},
  pages 1581--1584, 2010.

\bibitem{accenture-mm1}
K.~O'Brien, M.~Rybak, J.~Huang, A.~Stevens, M.~Fredriksz, M.~Chaberski,
  D.~Russell, L.~Castin, M.~Jou, N.~Gurrapadi, and M.~Bosch.
\newblock Accenture-mm1: A multimodal person recognition dataset.
\newblock In {\em Proceedings of the IEEE/CVF Winter Conference on Applications
  of Computer Vision (WACV) Workshops}, pages 112--122, January 2024.

\bibitem{open-set}
A.~Sapkota and T.~E. Boult.
\newblock Large scale unconstrained open set face database.
\newblock In {\em 2013 IEEE Sixth International Conference on Biometrics:
  Theory, Applications and Systems (BTAS)}, pages 1--8, 2013.

\bibitem{utk-lrhm}
Y.~Yao, B.~R. Abidi, N.~D. Kalka, N.~A. Schmid, and M.~A. Abidi.
\newblock Improving long range and high magnification face recognition:
  Database acquisition, evaluation, and enhancement.
\newblock {\em Computer Vision and Image Understanding}, 111(2):111--125, 2008.

\bibitem{Zhang_2020_CVPR}
F.~Zhang, X.~Zhu, H.~Dai, M.~Ye, and C.~Zhu.
\newblock Distribution-aware coordinate representation for human pose
  estimation.
\newblock In {\em IEEE/CVF Conference on Computer Vision and Pattern
  Recognition (CVPR)}, June 2020.

\bibitem{zhu2023gait}
H.~Zhu, Z.~Zheng, and R.~Nevatia.
\newblock Gait recognition using 3-d human body shape inference.
\newblock In {\em Proceedings of the IEEE/CVF Winter Conference on Applications
  of Computer Vision}, pages 909--918, 2023.

\bibitem{zou2020joint}
Y.~Zou, X.~Yang, Z.~Yu, B.~V. Kumar, and J.~Kautz.
\newblock Joint disentangling and adaptation for cross-domain person
  re-identification.
\newblock In {\em Computer Vision--ECCV 2020: 16th European Conference,
  Glasgow, UK, August 23--28, 2020, Proceedings, Part II 16}, pages 87--104.
  Springer, 2020.

\end{thebibliography}
}

\end{document}